\documentclass{article}

\PassOptionsToPackage{numbers,sort&compress}{natbib} 
\usepackage[preprint,sglblindworkshop]{neurips_2025}
\workshoptitle{ML$\times$OR Workshop 2025}

\usepackage[utf8]{inputenc}
\usepackage[T1]{fontenc}
\usepackage{microtype}
\usepackage{hyperref}
\usepackage{url}
\usepackage{xcolor}
\usepackage{booktabs}
\usepackage{amsmath, amssymb}
\usepackage{graphicx}

\title{Mechanistic Interpretability for Neural TSP Solvers}
\author{
  Reuben Narad\textsuperscript{1} \quad
  L\'eonard Boussioux\textsuperscript{1,2,*} \quad
  Michael Wagner\textsuperscript{1,*} \\
  \textsuperscript{1}University of Washington, Michael G. Foster School of Business \\
  \textsuperscript{2}University of Washington, Paul G. Allen School of Computer Science \& Engineering
  \textsuperscript{*}Equal supervision\\
  \texttt{\{rnarad, leobix, mrwagner\}@uw.edu}
}

\begin{document}

\maketitle

\begin{abstract}

Neural networks have advanced combinatorial optimization, with Transformer-based solvers achieving near-optimal solutions on the Traveling Salesman Problem (TSP) in milliseconds. However, these models operate as black boxes, providing no insight into the geometric patterns they learn or the heuristics they employ during tour construction. We address this opacity by applying sparse autoencoders (SAEs), a mechanistic interpretability technique, to a Transformer-based TSP solver, representing the first application of activation-based interpretability methods to operations research models. We train a pointer network with reinforcement learning on 100-node instances, then fit an SAE to the encoder's residual stream to discover an overcomplete dictionary of interpretable features. Our analysis reveals that the solver naturally develops features mirroring fundamental TSP concepts: boundary detectors that activate on convex-hull nodes, cluster-sensitive features responding to locally dense regions, and separator features encoding geometric partitions. These findings provide the first model-internal account of what neural TSP solvers compute before node selection, demonstrate that geometric structure emerges without explicit supervision, and suggest pathways toward transparent hybrid systems that combine neural efficiency with algorithmic interpretability. Interactive feature explorer: \url{https://reubennarad.github.io/TSP_interp/}.

\end{abstract}

\section{Introduction}

The Traveling Salesman Problem (TSP) is a canonical combinatorial optimization task with direct applications in vehicle routing, logistics, manufacturing, and network design. Although exact solvers, such as Concorde \citep{Applegate2006Concorde}, can certify optimality for instances up to tens of thousands of nodes, their exponential worst-case complexity and runtime growth motivate the development of fast approximation methods. Recent neural solvers, notably pointer networks and Transformer variants trained with reinforcement learning (RL), now construct high-quality tours rapidly once trained \citep{Kool2019,Berto2024RL4CO}. For example, on random Euclidean TSP-50, the attention model of \citep{Kool2019} achieves a 1.4\% optimality gap on its own, and 0.07\% with additional augmentations.  However, these models operate as black boxes: practitioners cannot identify \textit{which} geometric patterns the encoder represents, \textit{why} the decoder selects specific nodes during tour construction, or \textit{when} the policy might fail on out-of-distribution instances. 

This opacity poses practical barriers to industrial adoption. When a learned solver produces an unexpectedly poor tour, operators need to diagnose whether the failure stems from extrapolation beyond the training distribution, spurious correlations with irrelevant geometry, or misapplication of otherwise-sound heuristics. Without internal visibility, debugging requires expensive trial-and-error. This lack of transparency is especially problematic as organizations increasingly pair neural methods with classical Operations Research (OR) algorithms, where trust, auditability, and regulatory compliance are essential \citep{Kool2019, Berto2024RL4CO, Joshi2019, Zhou2025RoutingReview, wasserkrug2025enhancing}. Understanding the internal mechanisms of neural solvers would enable practitioners to predict failure modes, guide model corrections, and build hybrid systems that combine neural speed with algorithmic guarantees.

We address this interpretability gap by applying \textit{sparse autoencoders} (SAEs) to a Transformer-based TSP solver, marking to the best of our knowledge the first application of mechanistic interpretability, a subfield focused on reverse-engineering neural networks into human-legible components, to operations research models. Mechanistic interpretability (MI) decomposes model behavior into \textit{features} (interpretable directions in activation space representing distinct concepts) and \textit{circuits} (causal pathways connecting features to outputs). Recent MI advances center on SAEs, which resolve the \textit{superposition hypothesis}: modern networks represent far more features than they have neurons by encoding them as sparse, overlapping directions in activation space, causing individual neurons to respond polysemantically to multiple unrelated concepts \citep{Elhage2022ToyModels,Gao2024}. SAEs address this polysemanticity by learning an overcomplete but sparse dictionary of these directions, effectively disentangling neuron activations \citep{Olsson2022,Gao2024} and enabling researchers to identify monosemantic features that activate on single, interpretable concepts. This technique has uncovered internal mechanisms such as edge detectors in vision models, grammatical structures in language models, and protein binding motifs in genomic models, enabling targeted interventions and the formation of causal hypotheses \citep{Bricken2023Monosemantic,Anthropic2024ScalingMono}. 
 
 Our approach follows four steps: (i) train a pointer-style Transformer policy with REINFORCE on 100-node Euclidean TSP instances, achieving near-optimal solution quality, as benchmarked against Concorde, to ensure interpretability is performed on competitive policies; (ii) collect encoder activations from the final residual stream, after the last Transformer block processes the full spatial context—across many inference rollouts; (iii) fit a top-k sparse autoencoder to this residual stream to learn an overcomplete sparse dictionary of feature directions; and (iv) analyze the discovered directions by overlaying feature activations on multiple instances, visualizing how each feature responds across different node configurations \citep{Elhage2021,Olsson2022,Gao2024}. Across runs, we repeatedly observe features aligned with intuitive Euclidean-TSP notions: boundary-like responses on convex-hull nodes, cluster-sensitive responses in dense regions, and separator-like responses along linear or curved partitions, echoing classic human and heuristic intuitions while providing, for the first time, a model-internal account of what the encoder computes before the decoder's pointer step.
 
 Our contributions provide: (1) a reproducible pipeline for attaching SAEs to encoder-decoder optimization models, including hyperparameter configurations and training protocols; (2) empirical evidence that a competitive RL-trained TSP policy ($\approx$ 99\% optimal) develops human-interpretable geometric features without explicit inductive biases; (3) a taxonomy of recurring feature types (boundary, cluster, separator) discovered across multiple training runs; and (4) an interactive web-based feature explorer (\url{https://reubennarad.github.io/TSP_interp/}) enabling rapid hypothesis generation and qualitative analysis. While our analysis focuses on feature discovery in the encoder rather than causal circuit tracing to decoder outputs, we outline cross-layer transcoder architectures, activation patching experiments, and distributional robustness studies as natural extensions. These results represent an initial step toward transparent neural optimization systems where practitioners can inspect, validate, and refine learned heuristics alongside classical algorithms.

\section{Related Work}

The TSP, finding the shortest route to visit all cities in a list exactly once and return, is fundamental to combinatorial optimization with applications in logistics, manufacturing, and circuit design. Classical approaches rely either on computationally expensive exact algorithms or fast but suboptimal heuristics, hand-crafted from decades of mathematical insight. Motivated by this difficulty/accuracy tradeoff, there has been much interest in training deep learning models to solve these problems, both efficiently and near-optimally. One can think of these models as distilling a heuristic in the weights of the neural network.

\paragraph{Classical heuristics.}
Foundational constructive heuristics for constructing TSP solutions include nearest-neighbour and insertion methods, which trade solution quality for speed and simplicity. Local search schemes such as 2\hbox{-}opt and 3\hbox{-}opt perform iterated edge swaps and remain core post-improvement tools; the Lin--Kernighan (LK) heuristic and its Helsgaun implementation (LKH) are among the most effective practical solvers for large instances \citep{Helsgaun2000, LinKernighan1973}. For the metric TSP, Christofides’ algorithm achieves a worst-case $3/2$ approximation ratio and anchors much of the approximation literature \citep{vanBevern2020}. Metaheuristics, such as the Ant Colony System (ACS), also yield strong tours via stochastic constructive search guided by pheromone trails \citep{DorigoGambardella1997}. Exact solvers like Concorde combine branch-and-cut with sophisticated cutting planes and remain the gold standard for optimality certificates \citep{Applegate2006Concorde}.

\paragraph{Learning-based approaches to the TSP.}
Early neural approaches framed TSP tour construction as sequence prediction using Pointer Networks trained either with supervision on Concorde tours or with policy-gradient RL \citep{Vinyals2015,Bello2017NCO}. Transformer-based graph attention decoders trained with REINFORCE have become a widely adopted construction paradigm due to strong quality/speed trade-offs \citep{Kool2019}. Supervised GNN approaches predict edge/tour probabilities and can be paired with classical search for high-quality solutions \citep{Joshi2019}. Beyond autoregressive construction, there is growing work on \emph{non-autoregressive} (NAR) solvers that predict tours in parallel to reduce inference latency, e.g., an RL-trained NAR model (NAR4TSP) and diffusion-based NAR decoding that narrows the quality gap while retaining speed \citep{Xiao2024NAR4TSP,Wang2025DEITSP}. Complementing construction, \emph{learned improvement} replaces or augments local search: policies trained to perform 2\hbox{-}opt moves can iteratively refine tours and approach near-optimal quality faster than prior learned methods \citep{Costa2020TwoOpt}. Hybrid algorithms integrate learning signals into state-of-the-art heuristics, notably NeuroLKH, which learns edge scores and node penalties to guide LKH and improves generalization across sizes \citep{Xin2021NeuroLKH}. Finally, generalization beyond training sizes remains a central challenge; a controlled study finds that architecture and training choices strongly affect zero-shot performance, with autoregressive decoding providing a useful inductive bias but at higher inference cost \citep{Joshi2021Rethinking}.

\paragraph{Interpretability and mechanistic interpretability.}
MI is the study of how a trained neural network works by reverse‑engineering it into human‑legible parts: \emph{features} (what is represented) and \emph{circuits} (how parts compose to compute). In transformers, this typically means identifying directions in the model's latent space in attention heads, multi-layer perceptron (MLP) neurons, and the residual stream, that are most significant and meaningful \citep{Elhage2021,Olsson2022}.

A central tool in recent MI work is the \emph{sparse autoencoder} (SAE): given a dataset of internal activations of the subject model, an SAE learns an overcomplete but sparse basis in which individual feature directions are encouraged to be monosemantic and therefore easier to describe, visualize, and test. Recent studies also introduce metrics and scaling laws for feature quality \citep{Gao2024,Bricken2023Monosemantic,Anthropic2024ScalingMono}. Complementary to activation‑based MI, weight‑centric approaches analyze the structure of parameters directly; for example, “combinatorial interpretability” explains computations via sign patterns in weight and bias matrices without inspecting activations \citep{Adler2025Combinatorial}. In this paper, we adopt the activation‑based approach: we attach an SAE to the encoder’s residual stream to uncover geometric feature directions and leave causal circuit tests (e.g., patching) and weight‑centric analysis for future work.

\section{Training a Neural TSP Solver}

While there is a large diversity in deep learning methods for solving the TSP, a popular approach introduced by \cite{Kool2019} consists of a Graph Transformer (GAT) trained using RL. 

\subsection{Neural Network Architecture}

\begin{figure}[th]\label{fig:encoder-decoder}
  \centering
  \makebox[\linewidth][c]{%
    \includegraphics[width=1.2\linewidth]{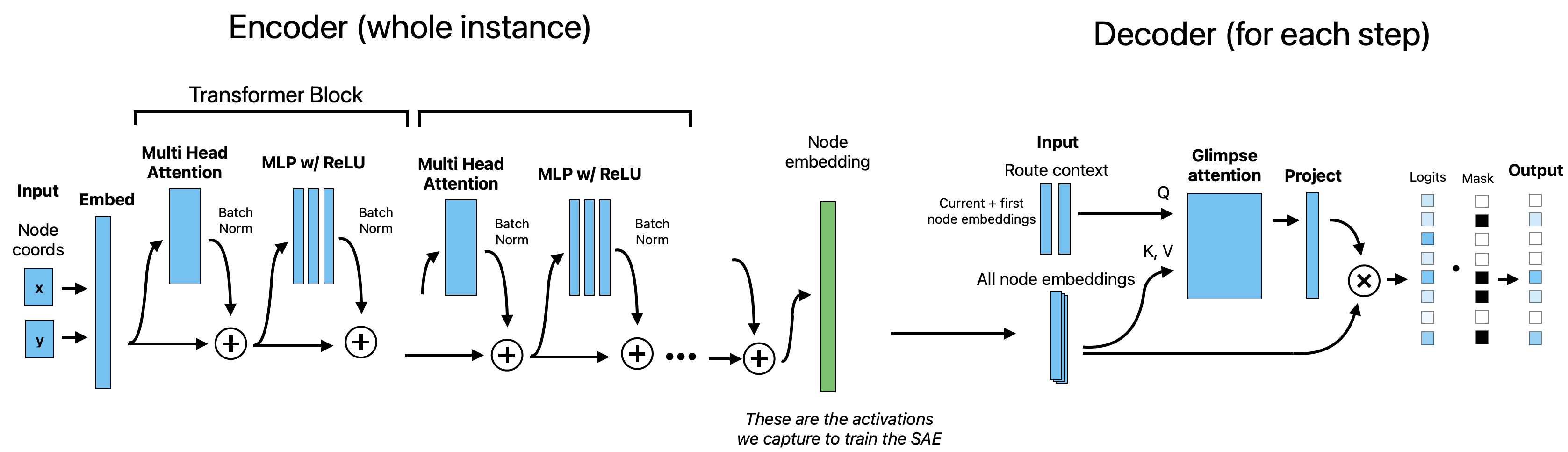}%
  }
  \caption{System architecture of the TSP solver transformer model, adopting the architecture from \cite{Kool2019}.}
\end{figure}

The Graph Transformer is implemented as an encoder-decoder (see Figure \ref{fig:encoder-decoder}). The encoder takes as input the list of nodes with their $(x,y)$ coordinates, together with one-hot features that indicate whether a node is the current node, the terminal node, and whether it has already been visited. It performs a single forward pass for the whole trajectory, producing embeddings that represent the entire set of nodes jointly. The decoder then consumes the embedding of the whole graph along with the embedding of the current node, and outputs logits over the input nodes. We construct the tour autoregressively by greedily selecting, at each step, the node with the highest output logit. For the network, we use RL4CO’s \cite{Berto2024RL4CO} implementation.

\setlength{\textfloatsep}{10pt plus 1pt minus 2pt} 
\setlength{\intextsep}{7pt plus 1pt minus 2pt}    

\subsection{Training with RL}
\begin{figure}[!h]
  \centering
  \begin{minipage}{0.3\linewidth}\centering
    \includegraphics[width=\linewidth]{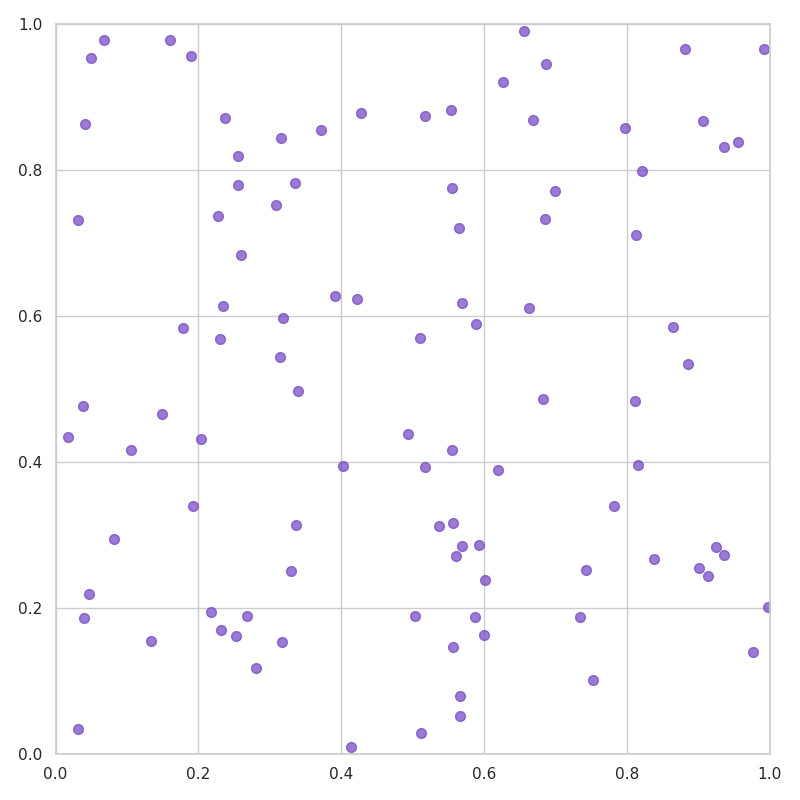}\\
    \emph{Uniform}
  \end{minipage}\hfill
  \begin{minipage}{0.3\linewidth}\centering
    \includegraphics[width=\linewidth]{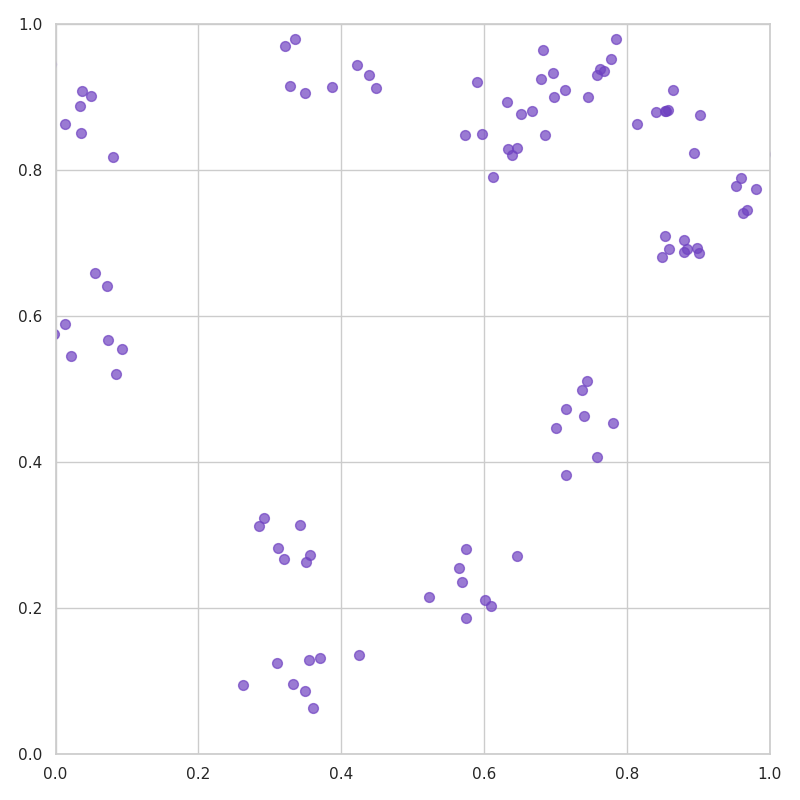}\\
    \emph{Clusters}
  \end{minipage}\hfill
  \begin{minipage}{0.3\linewidth}\centering
    \includegraphics[width=\linewidth]{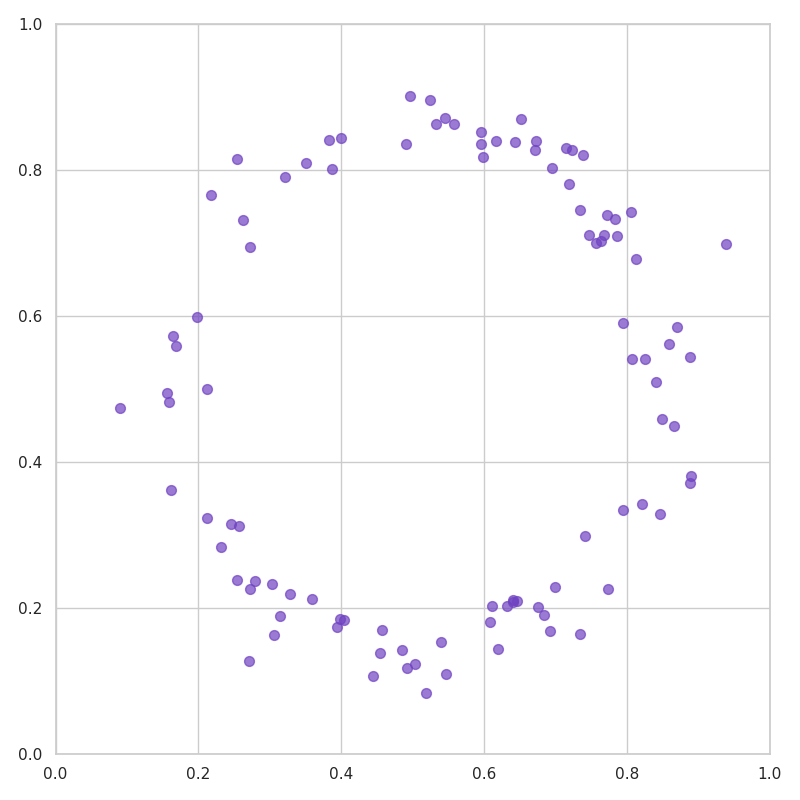}\\
    \emph{Ring}
  \end{minipage}
  \caption{Example TSP instances drawn from different distributions.}\label{fig:distrib}
\end{figure}

Because the TSP is NP-hard, obtaining supervised-learning labels at scale would be slow and computationally expensive. RL provides an alternative: we can roll out an unbounded number of trajectories and obtain reward automatically as the negative completed tour length \cite{Joshi2019}. In our setup, the environment draws nodes from a uniform distribution on the unit square; the same approach can learn other distributions as well (see Figure \ref{fig:distrib}).

We train with a modified version of the REINFORCE algorithm \cite{Williams1992}, using a clipped objective, noting that other RL algorithms (e.g., PPO) have also been shown to be effective. Each training run begins with a warm-up phase of 1{,}000 greedy rollouts to initialise the baseline and stabilise early learning. Afterwards, node selection during rollouts is sampled from a temperature-controlled softmax, and once REINFORCE produces gradients, we update parameters with the Adam optimizer.

\section{Interpretability with Sparse Autoencoders}
\subsection{Interpretability Goal}
Understanding what the neural TSP solver has learned requires dissecting its latent space in a human-interpretable way.  We adopt the language of \cite{Elhage2021}, where ``features'' are important directions in the model's latent space that behave like variables: each has a value (its activation) on every forward pass. ``Circuits,'' on the other hand, are relationships between these features that behave like functions. We focus on finding ``features.'' 

We look for these features in neuron \emph{activations} rather than raw weights because activations are input–conditional and expose the instance–specific computations actually performed by the policy, whereas weights are input–agnostic and can entangle many behaviors via superposition. Sparse autoencoders operate directly on these activations to learn an overcomplete but \emph{sparse} basis, yielding features that are far less polysemantic and thus easier to describe and test. We note that \emph{weight}-centric approaches exist, e.g., “combinatorial interpretability,” which analyses the combinatorial structure of weight and bias signs to explain computation without examining activations, see \cite{Adler2025Combinatorial} (static analysis of weight matrices, no activations) for a recent example.

\subsection{SAE Background}
A common obstacle in interpreting neural networks is \emph{polysemanticity,} a phenomenon where individual neurons encode multiple concepts \cite{Olsson2022}. In language models, for example, this could be a neuron that activates on both French negations and HTML tags, thereby confounding interpretation. A popular tool for ``disentangling'' the neurons is the Sparse Autoencoder (SAE), a secondary ML model trained on the activations of the model being interpreted. SAEs learn an \emph{over-complete} and \emph{sparse} representation of the activations \citep{Olsson2022}.

\begin{figure}[th]\label{fig:sae}
  \centering
  \includegraphics[width=0.5 \textwidth]{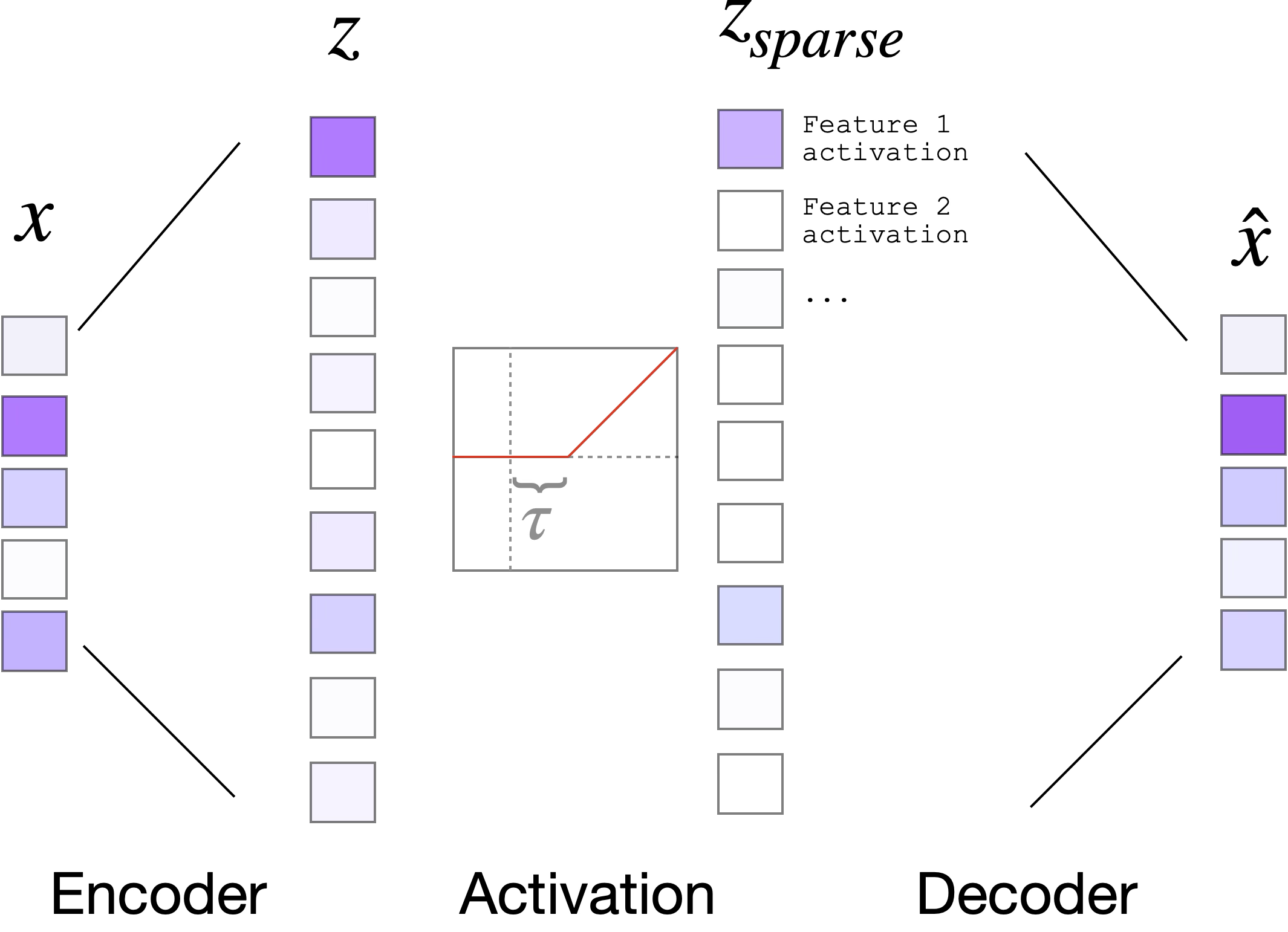}
  \caption{The SAE learns an overcomplete and sparse representation of neuron activations of the TSP solver model}
\end{figure}

The SAE architecture consists of three components: an encoder, an activation function, and a decoder. Its training objective, as an autoencoder, is to reconstruct the input after passing it through its encoder's latent space. As a compression task, autoencoders typically have a smaller-dimensional latent space than the input. The key difference with \emph{Sparse Autoencoders} is that the latent space is higher-dimensional, with the additional sparsity constraint (see Figure \ref{fig:sae}).

Mathematically, the SAE forward pass consists of three steps. Given a node embedding $x \in \mathbb{R}^d$, the \emph{encoder} first projects to the latent space:
\[
z = xW_{\text{enc}}^{\top} + b,
\]
where $W_{\text{enc}} \in \mathbb{R}^{n \times d}$ contains the learned feature directions and $b \in \mathbb{R}^n$ is a bias vector. Next, the \emph{sparsification} step applies top-$k$ activation:
\[
z_{\text{sparse}} = \text{TopK}(z) = \text{ReLU}(z - \tau),
\]
where $\tau$ is the $k$-th largest value in $z$, effectively zeroing all but the $k$ largest activations, creating the sparse bottleneck. Finally, the \emph{decoder} reconstructs the input:
\[
\hat{x} = z_{\text{sparse}}^{\top} W_{\text{dec}} + b_{\text{dec}},
\]
where $W_{\text{dec}} \in \mathbb{R}^{n \times d}$ maps the sparse latent representation back to the original space.

Due to the sparsity constraint, the discovered features are less likely to be polysemantic and more interpretable to humans. 

\subsection{SAE Training Details}

We attach the SAE to the encoder's final residual stream—after all Transformer blocks have processed the input—to capture the richest spatial representation before autoregressive decoding begins. Our architecture follows the top-$k$ SAE framework of \cite{Gao2024}: at each forward pass, we retain only the $k$ largest activations per token while zeroing the remainder, imposing an effective $\ell_0$ sparsity constraint while preserving gradient flow through the differentiable top-$k$ operation.

Three key hyperparameters govern the SAE's behavior: (i) the \emph{expansion factor}, which determines the width of the SAE as a ratio of the latent dimension to the encoder's embedding dimension; (ii) the \emph{$k$-ratio}, which controls sparsity by setting what fraction of latent features can activate per token; and (iii) the \emph{$\ell_1$ coefficient}, which weights the sparsity penalty against reconstruction fidelity in the loss function. These parameters induce competing objectives: increasing the $k$-ratio improves reconstruction quality by allowing more features to fire simultaneously, but reduces both sparsity and feature specialization; conversely, increasing the $\ell_1$ coefficient promotes sparser representations but degrades reconstruction accuracy.

To identify a suitable operating point, we conducted a grid search over expansion factor $\in \{1, 4, 16\}$, $k$-ratio $\in \{0.01, 0.10\}$, and $\ell_1$ coefficient $\in \{10^{-4}, 10^{-3}, 10^{-2}, 10^{-1}\}$. For each configuration, we trained a separate SAE on encoder residuals from 100,000 TSP inference passes, holding the optimizer (Adam), batch size, and learning rate schedule fixed across all runs. We selected the configuration $(\text{expansion factor} = 4, k\text{-ratio} = 0.1, \ell_1\text{-coefficient} = 0.001)$ based on qualitative assessment of feature interpretability and reconstruction error, though we acknowledge this selection procedure warrants more systematic evaluation in future work. The resulting SAE achieves near-perfect reconstruction of residual activations while maintaining the target sparsity level throughout training. 

We collect the SAE's training data by running inference on the TSP model for 100,000 instances, collecting the graph embedding for each one. Crucially, these instances were drawn from the same distribution as the TSP solver's training data. Using our tuned hyperparameter configuration, the resulting dictionary reconstructs residual activations near-perfectly, with increasing sparsity as training progresses.

\section{Feature Analysis}
\subsection{Feature Activation Mechanics}
After identifying a set of features with the SAE, we aim to understand how each feature corresponds to distinct TSP attributes. A ``feature activation'' is simply the activation value of a specific neuron (feature) in the SAE's latent space during a forward pass. For a given node embedding $x$, we compute the SAE's sparse latent representation $z_{\text{sparse}}$ using the encoder and top-$k$ sparsification described above. The activation of feature $i$ is then just $z_{\text{sparse}}[i]$, the $i$-th component of this sparse vector.

Since each TSP instance contains multiple nodes, feature $i$ produces a collection of activations $\{z_{\text{sparse}}[j,i]\}_{j=1}^{N}$ across all $N$ nodes. To summarize how strongly a feature responds to an entire instance, we compute the mean activation, which helps us sort features by relevance for different types of TSP instances:
\[
\mu_i \;=\; \tfrac{1}{N}\sum_{j=1}^{N} z_{\text{sparse}}[j,i].
\]

\subsection{Visualizing a Feature}
To see what a feature is doing, we overlay ten 100-node instances from the same distribution in a single $x$-$y$ plot. Each point's color encodes the corresponding value of $z_{\text{sparse}}[j,i]$ (dark = 0, bright = high activation), while marker shape distinguishes which of the ten instances the node belongs to. Because activations are node-wise, this composite heatmap allows us spot geometric or combinatorial regularities at a glance, e.g., gradients that track tour direction, clusters of high-activation nodes near dense regions, or symmetry patterns that correlate with particular edge layouts.

\subsection{Discussion}
For demonstration, we trained an SAE on a model trained on uniform distributions. As shown in Figure~\ref{fig:sae_features} in Appendix A, we found many recurring themes among the features. These feature categories suggest that the SAE successfully disentangles different aspects of spatial reasoning that are important for TSP solving, ranging from boundary detection to spatial clustering and geometric separation. 
{Our next goal is to recover circuits: use cross-layer transcoders to map features in the encoder residual stream to node-selection behavior in the decoder, then confirm causality with patching and head/MLP ablations. 
Ultimately, we aim to distill these circuits into transparent, reusable OR primitives, i.e., to understand what these models are really doing.}

\section{Limitations and Future Work}

This study is deliberately exploratory. While we uncover semantically coherent, geometry-aligned directions in the encoder via an SAE, our analysis is primarily correlational and limited in scope. We outline the main constraints and next steps.

\textbf{From correlation to computation (across features and layers).}
Our overlays and qualitative examples do not yet establish that discovered features \emph{cause} specific node selections or tour improvements, nor how they compose across layers to drive pointer logits. A more causal story requires tracing computations \emph{across features and layers}: (i) feature\(\rightarrow\)logit analyses (do particular feature activations predict pointer probabilities after controlling for coordinates?); (ii) activation patching/steering and selective ablations (does toggling a feature predictably shift next-node probabilities and tour length?); and (iii) mapping feature usage into attention heads and MLPs. A key next step is a \emph{cross-layer transcoder} that learns how encoder features transform into decoder-internal representations, enabling hypothesis tests about circuits that connect feature groups to pointer decisions.

\textbf{Scope and external validity.}
Most of our analyses use a transformer trained at \(N{=}100\) on instances drawn i.i.d.\ from a uniform unit square, with only illustrative examples from other geometries. This limits claims about generalisation across sizes and distributions. We plan systematic evaluations across sizes (\(N\in\{50,100,200,500\}\)) and geometries (clustered Gaussians, rings, road-network–like layouts), including train/test distribution mismatch, to quantify which features persist, shift, or disappear.

\textbf{Methodological dependence on the SAE and the underlying policy.}

Our feature discoveries depend on multiple interconnected design choices. First, the SAE itself introduces several architectural degrees of freedom: which layer to attach to (tap layer), how much to expand the dictionary (expansion factor), how aggressive the sparsity constraint is (sparsity level), and algorithmic choices (optimizer, random seed). Second, our activation-based approach exposes instance-conditional computations—revealing what the model does on specific inputs—but dictionary-learning objectives can inadvertently introduce artifacts or simply mirror distributional properties of the training data rather than capturing genuine algorithmic structure.

Critically, the features we discover also reflect how the \emph{underlying policy} was trained. Choices in the RL training procedure, reward shaping, decoding temperature, baseline initialization, warm-up schedules, and exploration noise, all shape the residual-stream geometry that the SAE ultimately learns from. Different training configurations may yield different internal representations, even if the final tour quality is similar.
To address these concerns, future work will plan several robustness checks: (i) sweeping SAE hyperparameters and attachment points to assess feature stability; (ii) reporting quantitative feature-quality metrics (reconstruction error, sparsity, consistency across seeds); (iii) comparing against non-SAE decompositions (PCA, ICA, NMF) to verify that discovered structure is not merely an artifact of the SAE objective; and (iv) contrasting our activation-based findings with recent \emph{weight}-centric interpretability methods that analyze parameter sign structure directly without examining activations \citep{Adler2025Combinatorial}.

\textbf{Breadth of model classes.}
Our main experiments use an autoregressive construction policy. To assess whether similar geometric representations emerge more broadly, we intend to replicate the analysis for supervised GNN constructions, non-autoregressive decoders, learned-improvement policies (e.g., 2\hbox{-}opt–style policies), and hybrid methods (e.g., NeuroLKH). Convergent (or principled divergent) feature families across these settings would clarify which representations are characteristic of \emph{neural} TSP solving versus specific to an architecture or training regime.

\section{Conclusion and Future Work}

We introduced an activation-based, SAE-driven lens on a neural TSP solver and documented recurring, human-aligned geometric directions (boundary, cluster, separator) in the encoder of a strong RL-trained policy. This offers a first, model-internal account of what is computed before the pointer step and complements classical heuristics and recent learning-based solvers.

Looking ahead, our priorities are (i) moving from correlation to computation by tracing \emph{across features and layers} and building a \emph{cross-layer transcoder} to test causal paths from encoder features to decoder logits; (ii) systematic size/distribution studies to characterise which features persist under scaling and shift; (iii) robustness analyses for the SAE (hyperparameters, tap layers, seeds) and comparisons with non-SAE decompositions and weight-centric analyses; and (iv) breadth across model families, including non-autoregressive, learned-improvement, and hybrid (learning-guided LKH) approaches. These steps will turn qualitative observations into falsifiable circuit hypotheses and, ultimately, into tools for hybrid, trustworthy routing systems.





\clearpage
\bibliographystyle{plainnat}
\bibliography{references} 


\appendix
\section{Example visualizations of discovered SAE features}

\newlength{\overlaygap}
\setlength{\overlaygap}{0.4em} 

\newlength{\overlayw}
\setlength{\overlayw}{\dimexpr(\linewidth-2\overlaygap)/3\relax} 

\newcommand{\overlayblock}[5][4.2cm]{%
  \begin{minipage}{\linewidth}\centering
    {#2}\\[-0.2em]
    \vspace{.2cm}
    \includegraphics[width=\overlayw,clip]{#3}\hspace{\overlaygap}%
    \includegraphics[width=\overlayw,clip]{#4}\hspace{\overlaygap}%
    \includegraphics[width=\overlayw,clip]{#5}
  \end{minipage}\par\vspace{0.35em}%
}

\begin{figure}[th]
  \centering
  \overlayblock{Activates on the edge}
    {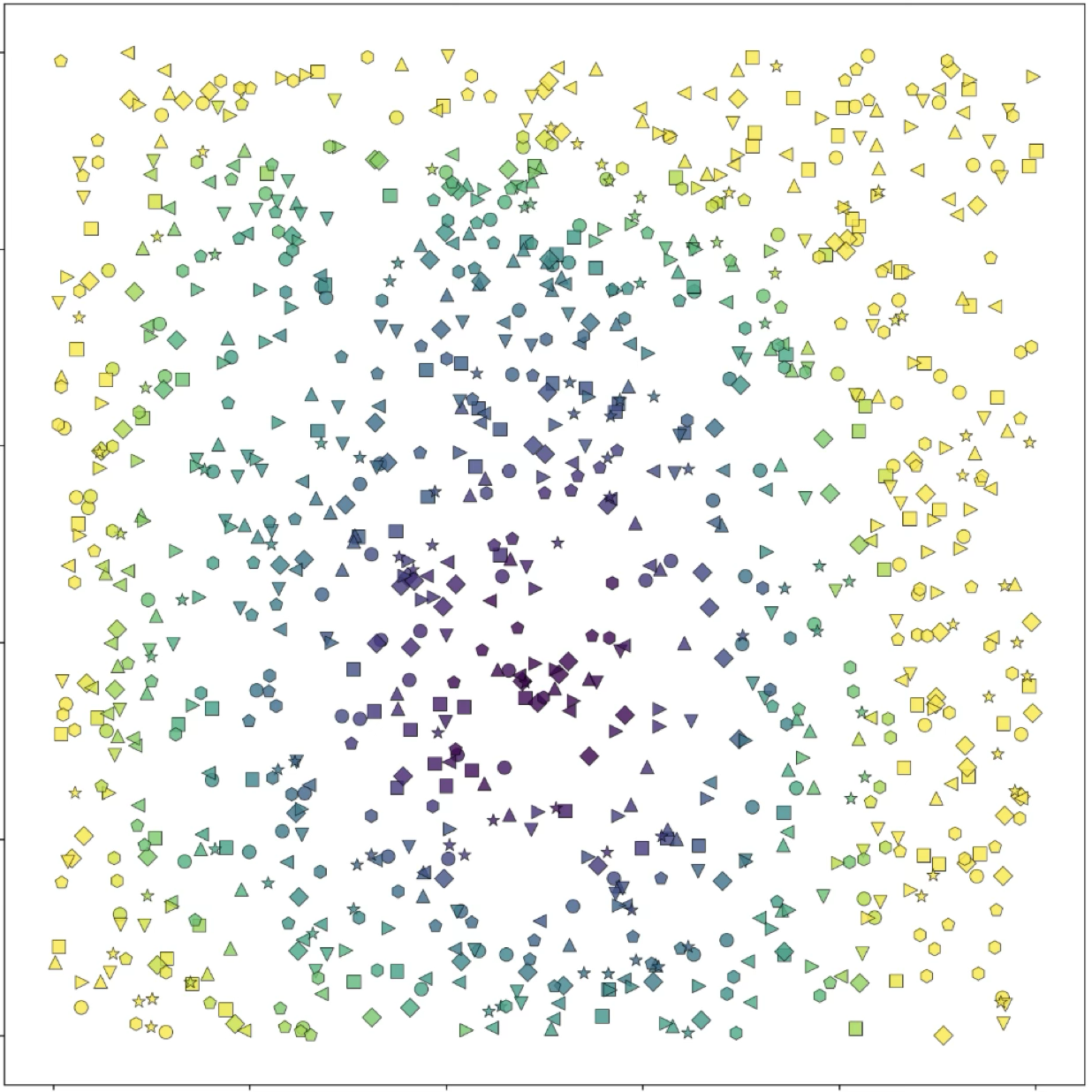}{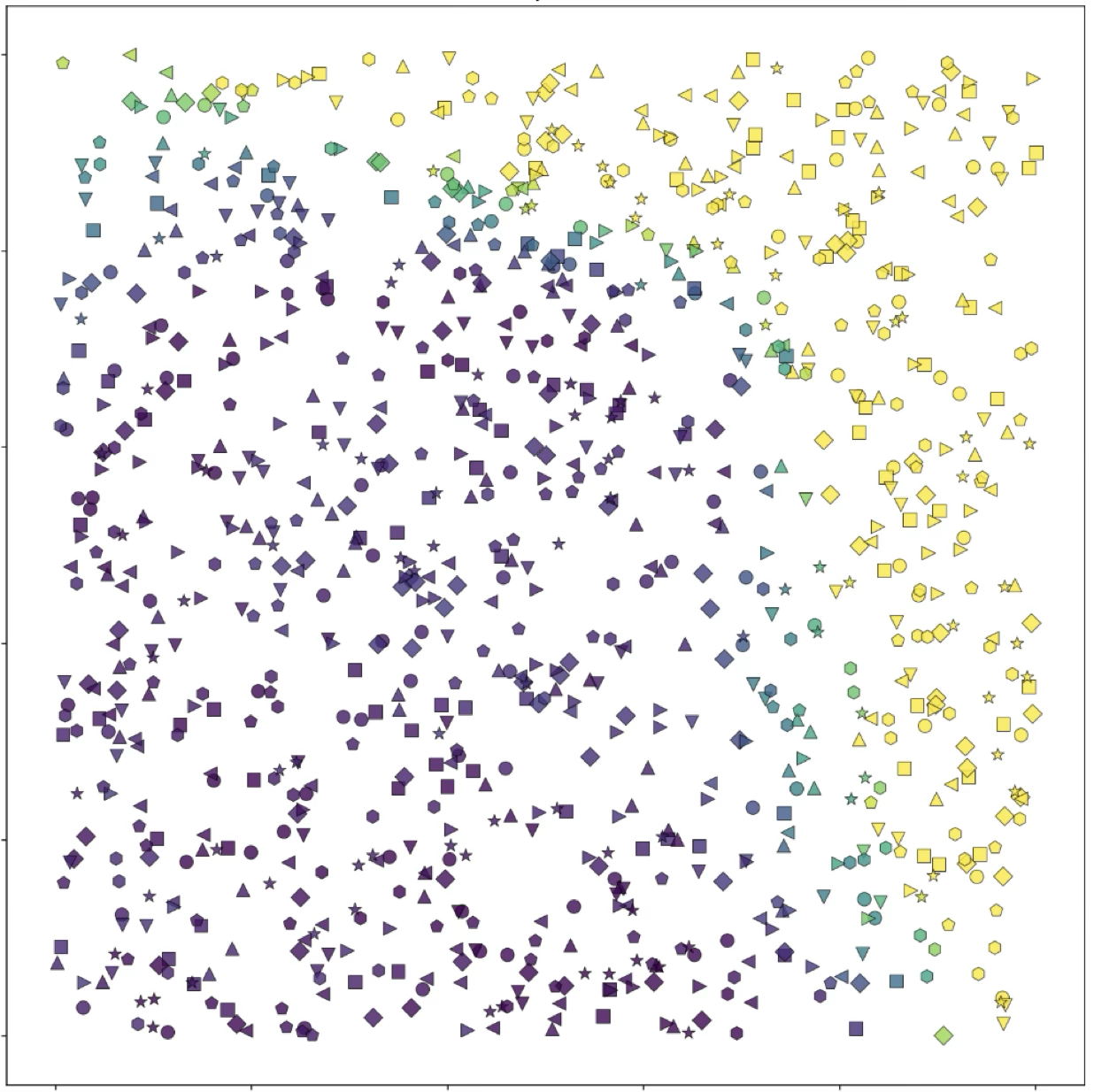}{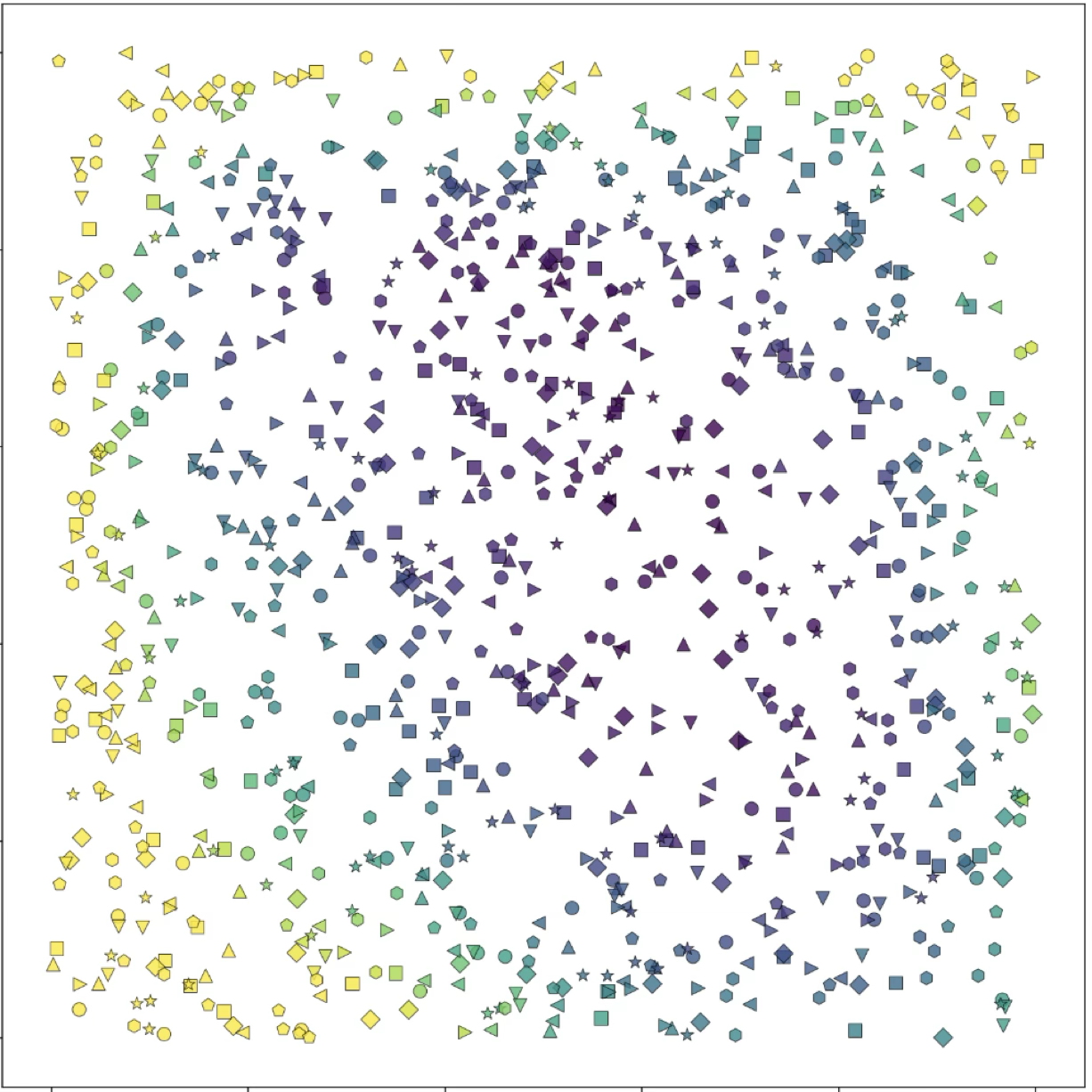}
  \vspace{.3cm}
  \overlayblock{Focuses on one spot}
    {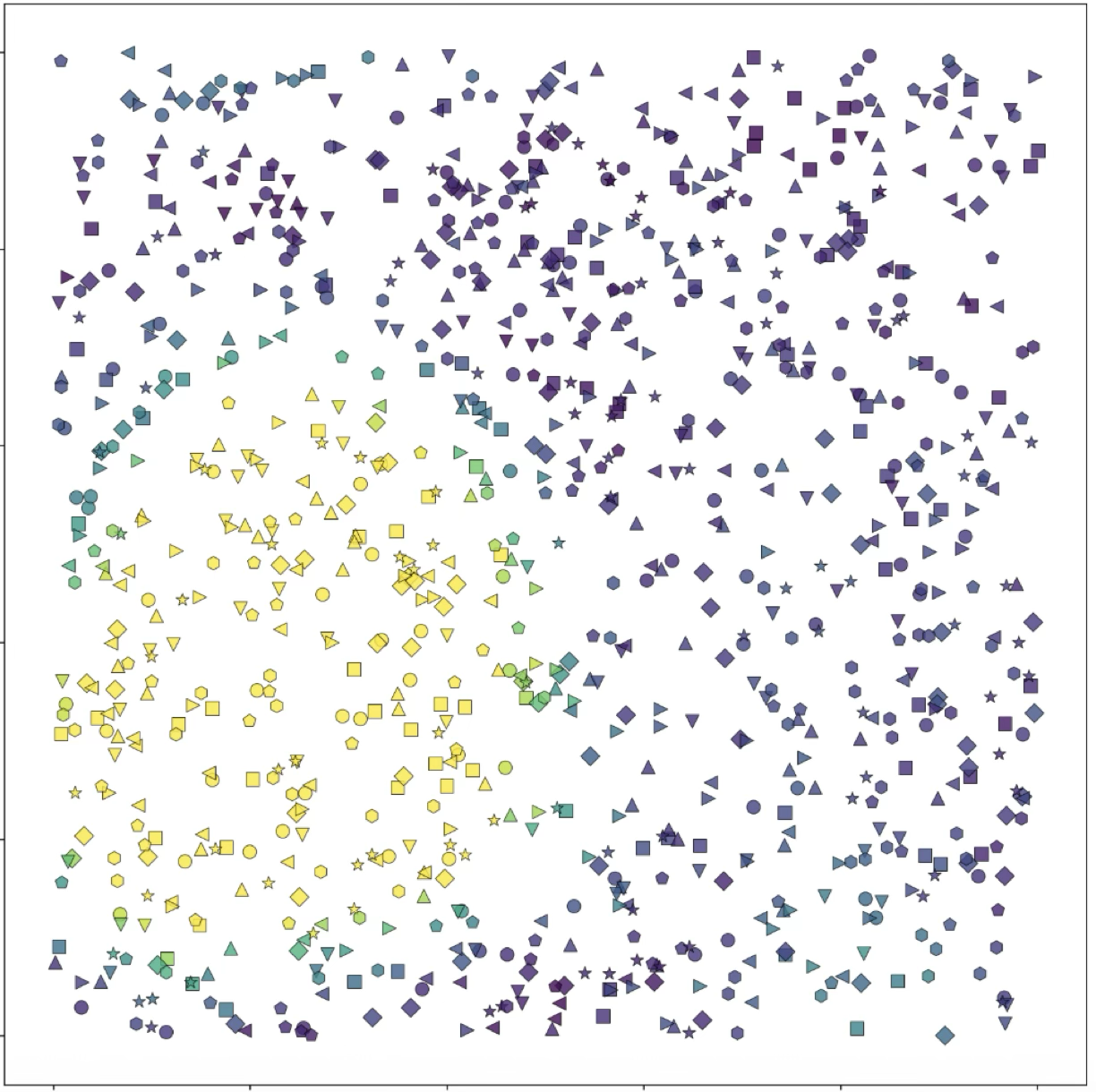}{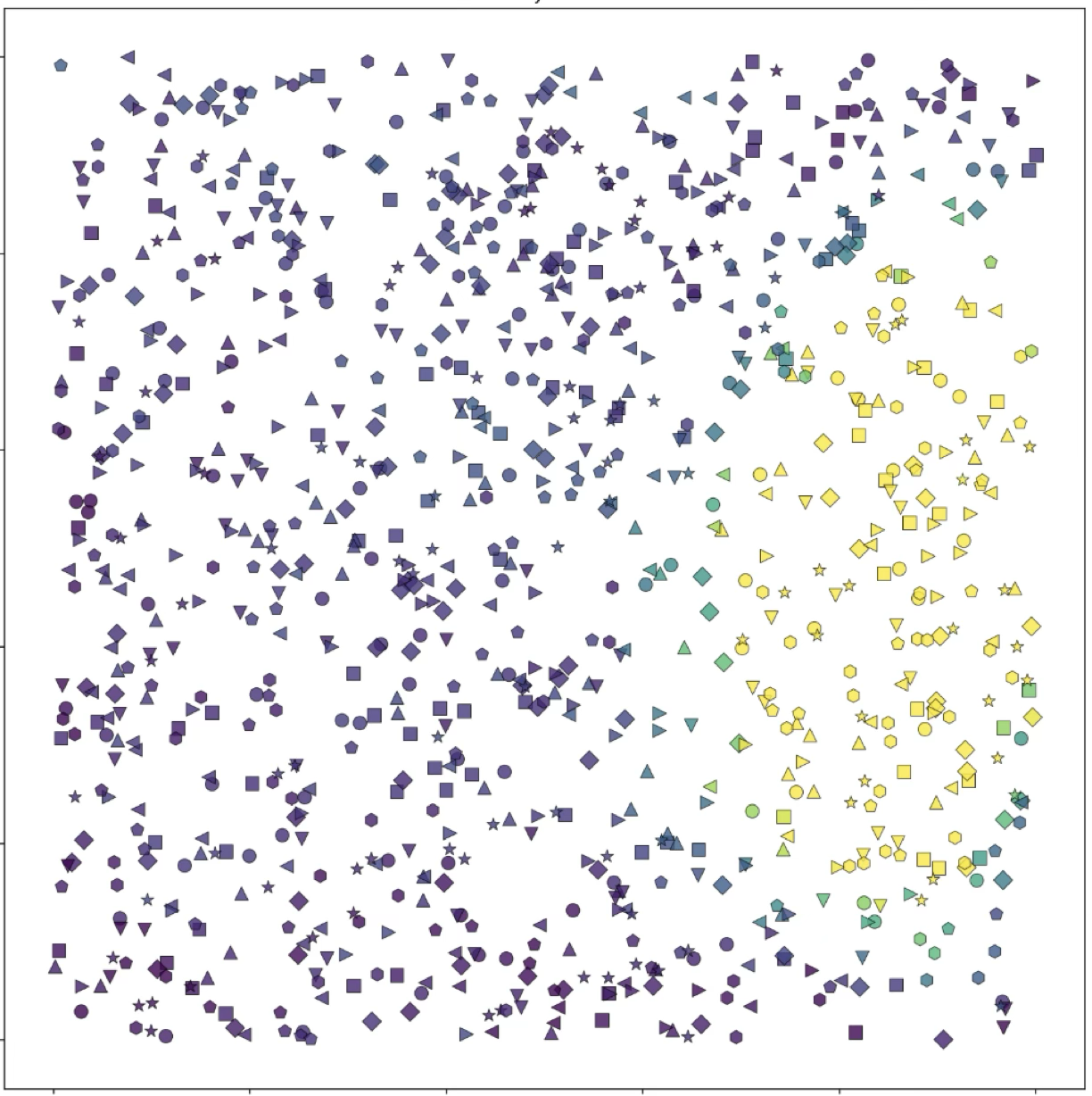}{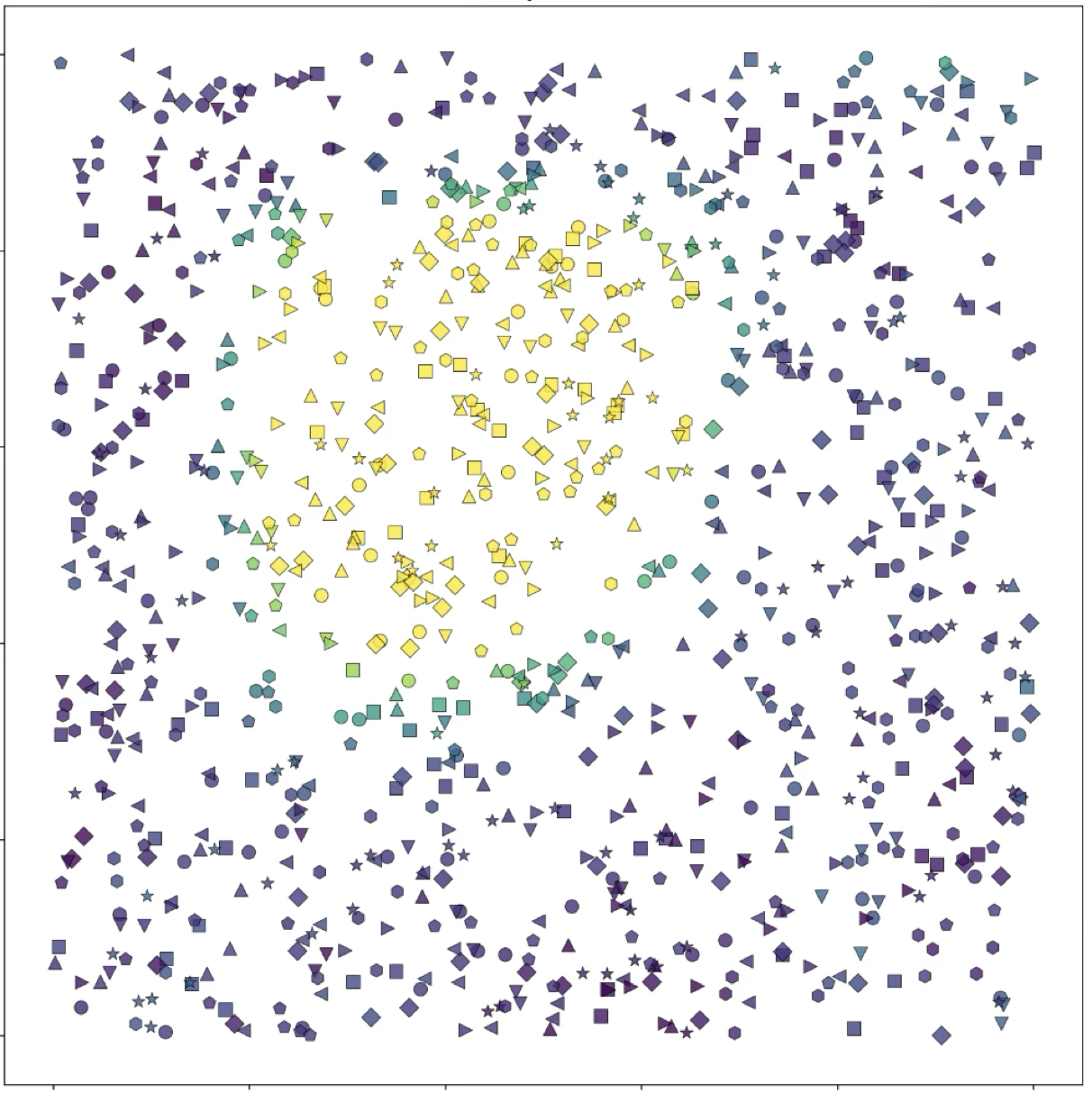}
  \vspace{.3cm}
  \overlayblock{Linearly separated}
    {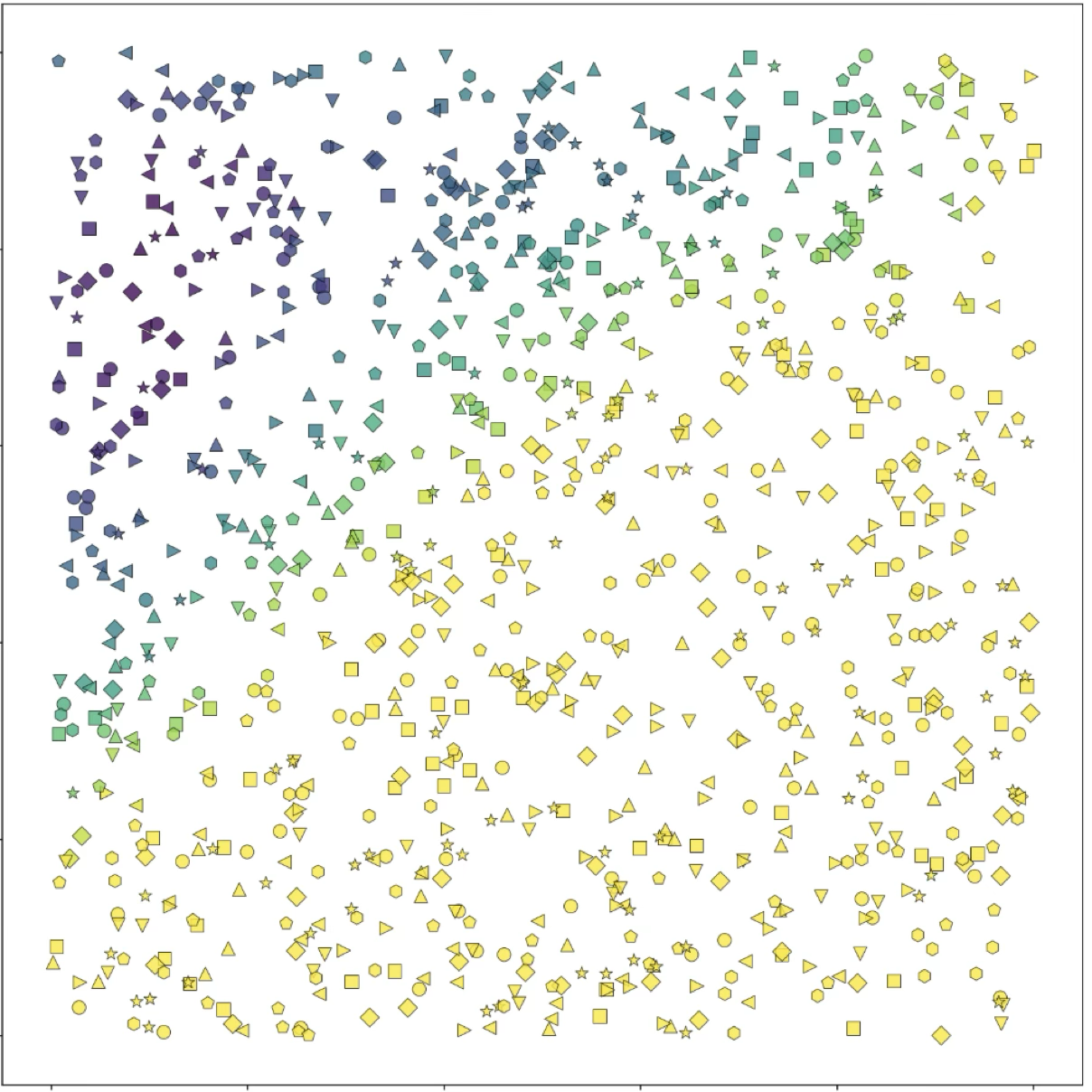}{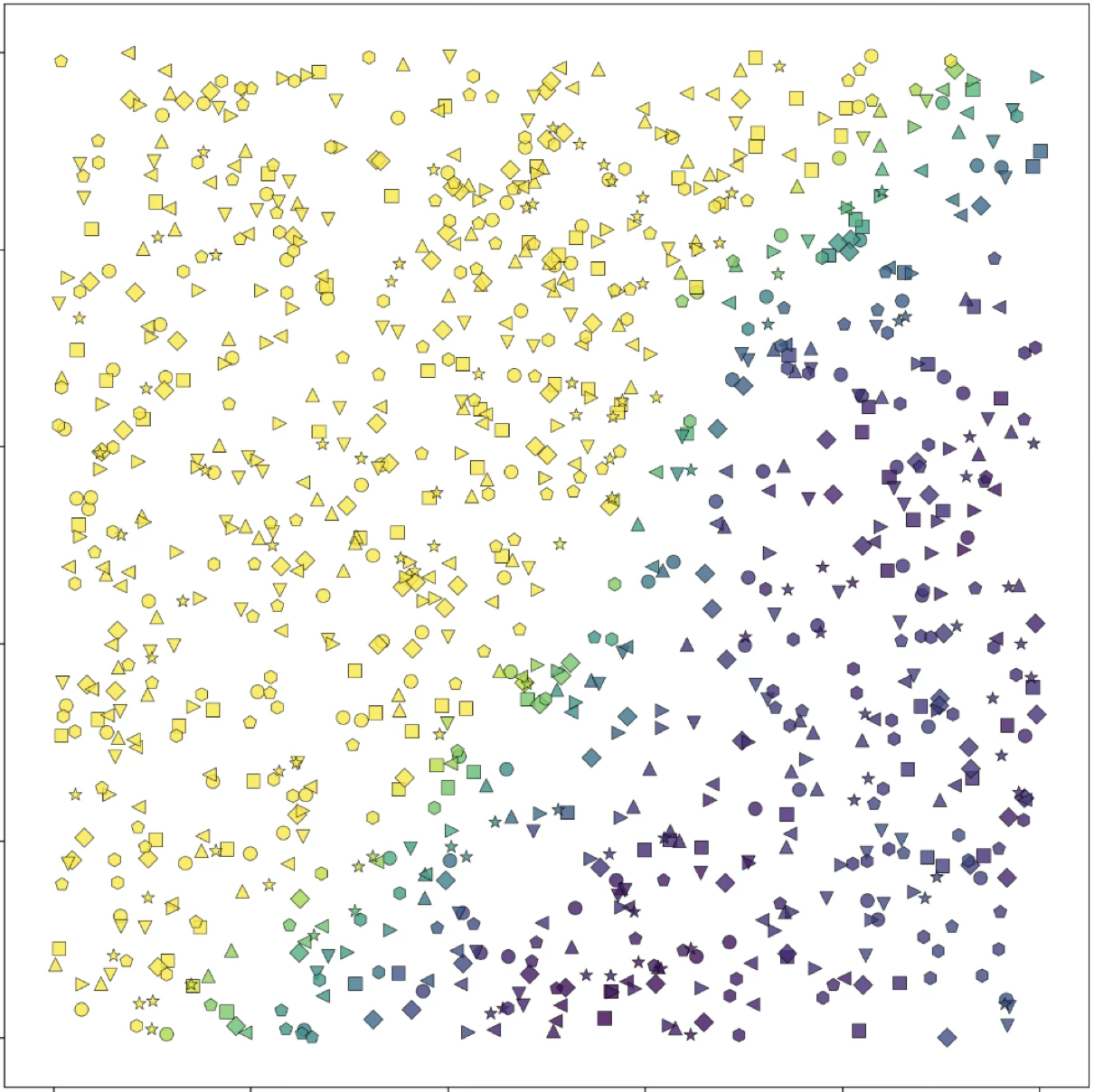}{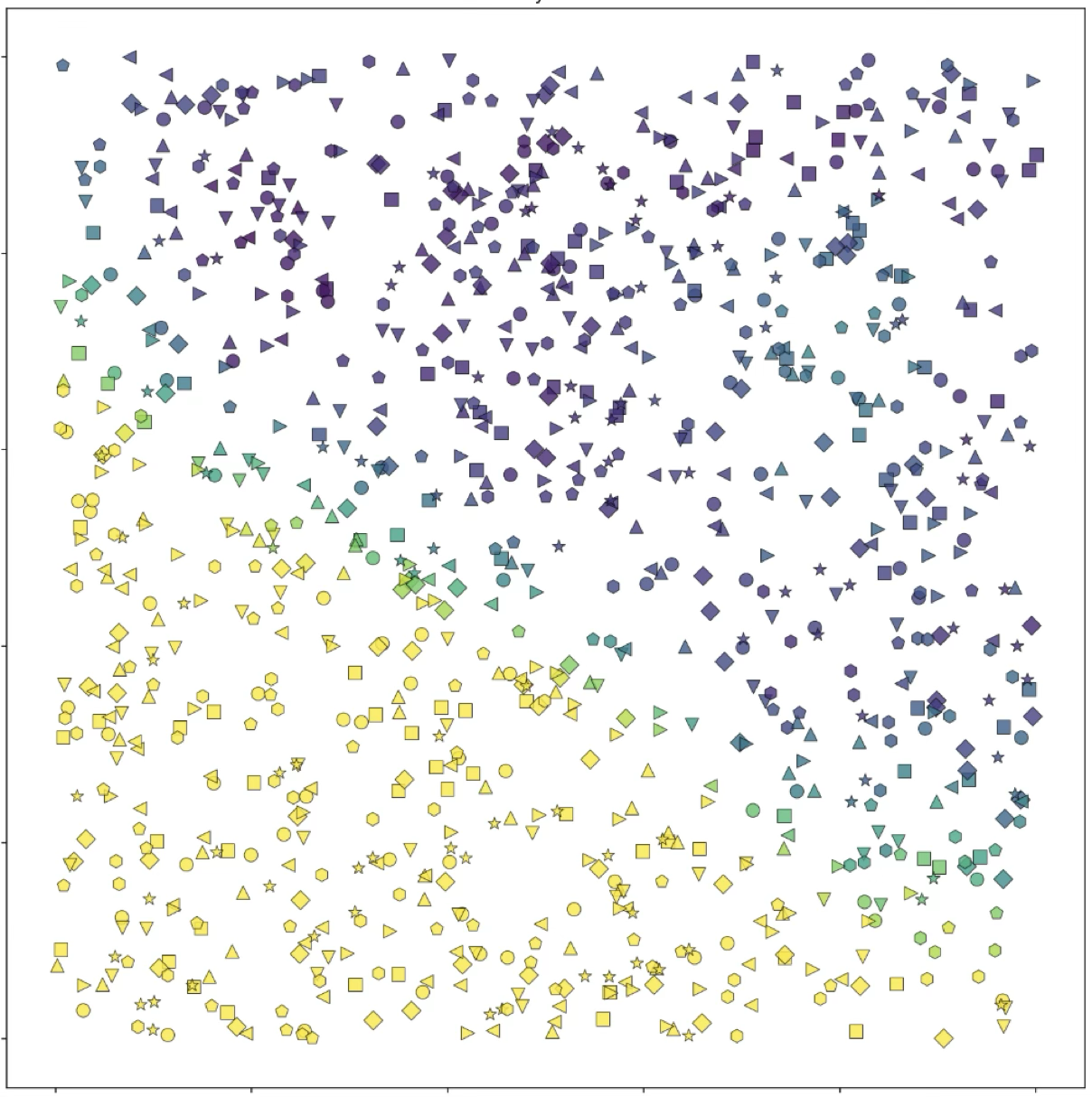}
  \vspace{.3cm}
  \overlayblock{Unclear}
    {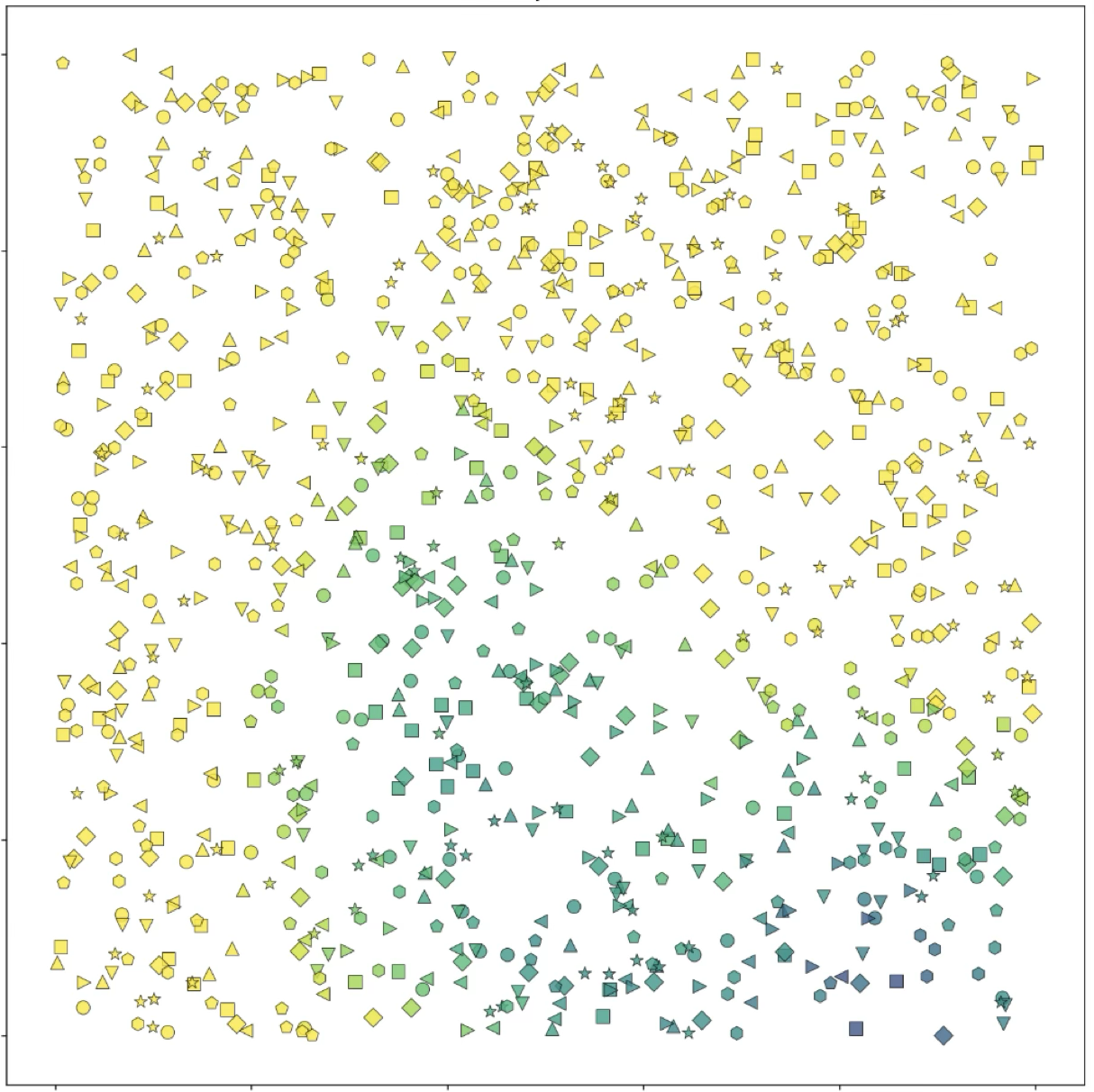}{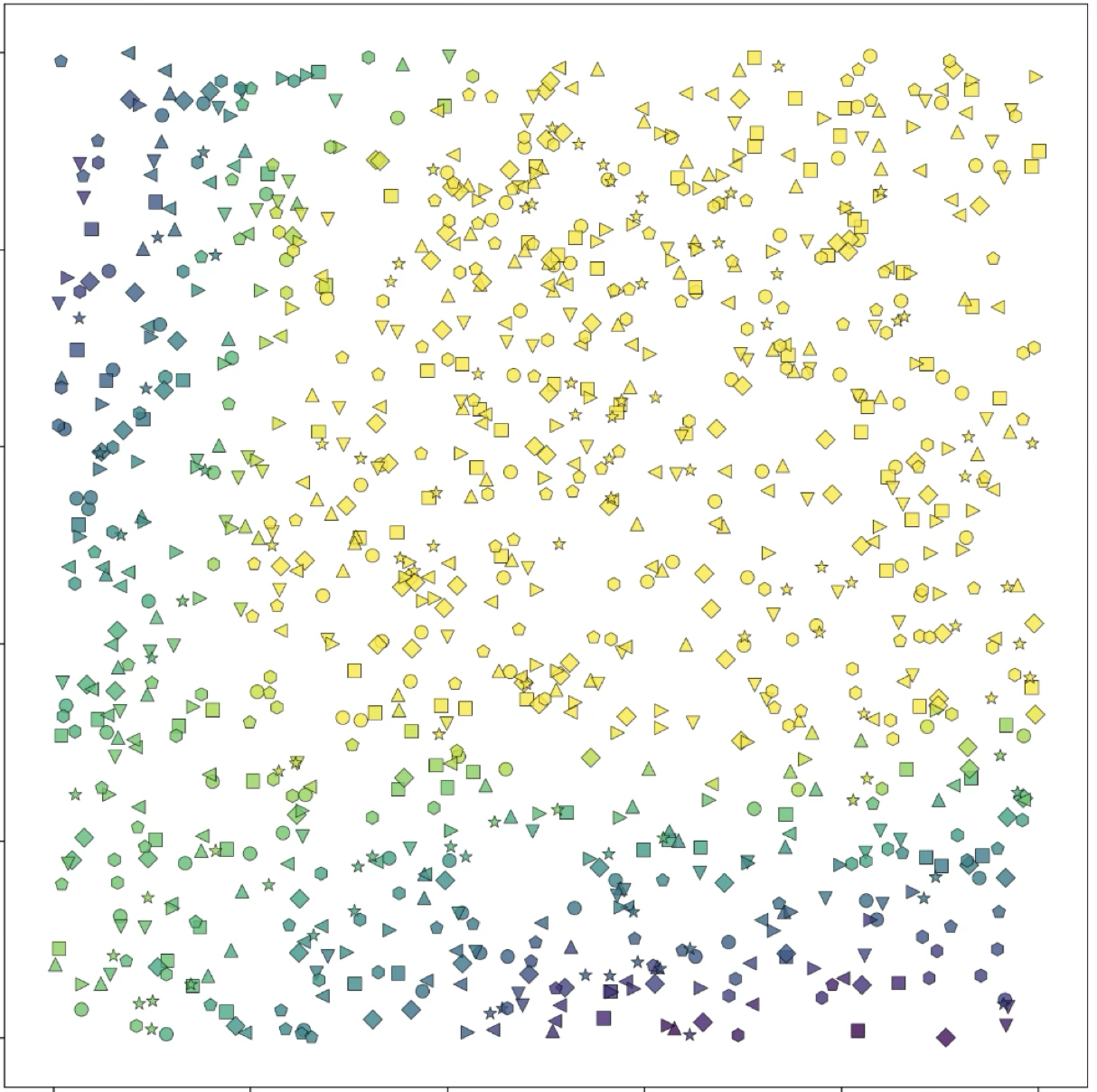}{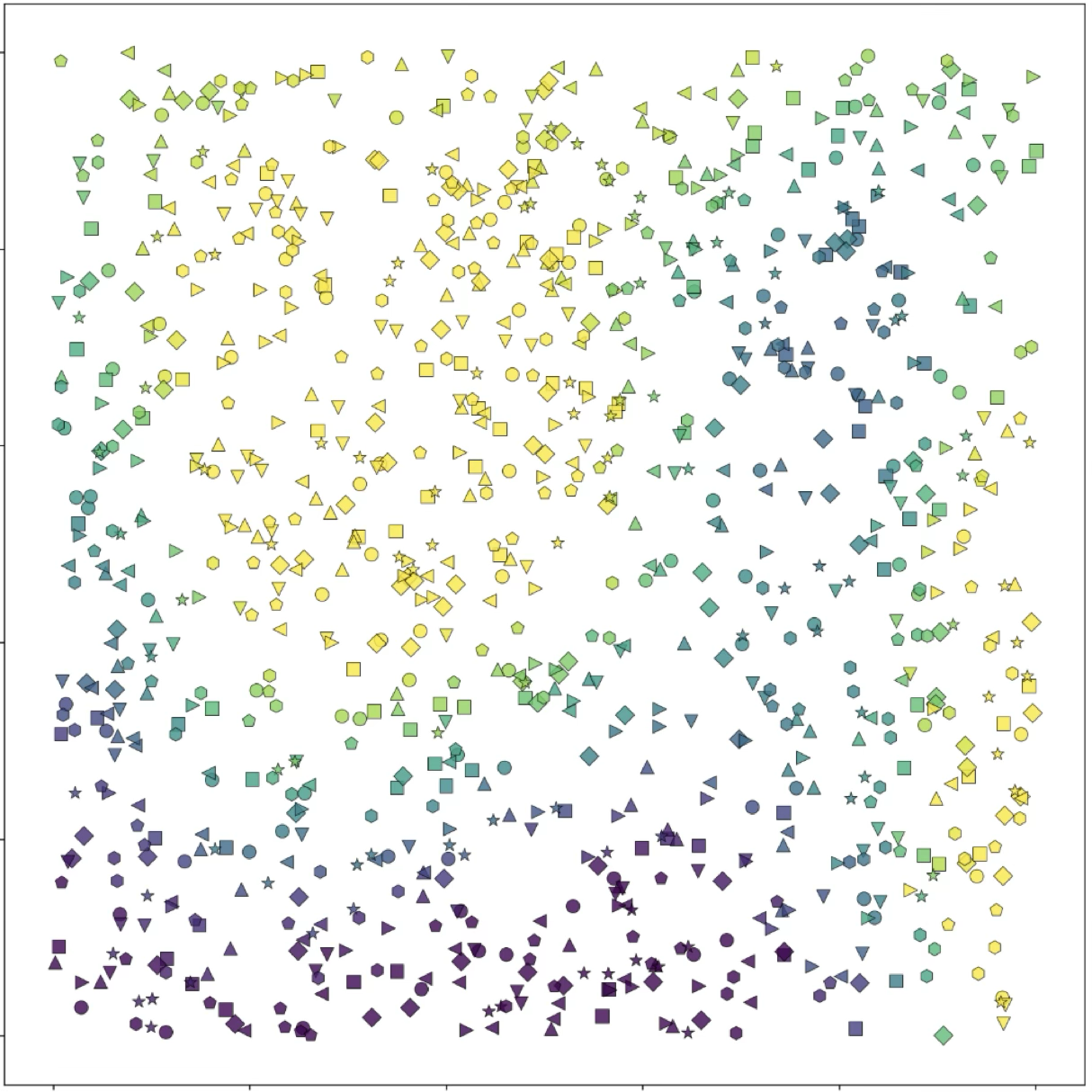}
\caption[SAE overlays]{Each panel is a visualization of a single SAE feature. It overlays nodes from 10 TSP instances, with the marker shape indicating the instance this node was drawn from. Color shows SAE activation, with purple = low and yellow = high.}
  \label{fig:sae_features}
\end{figure}

\end{document}